\documentclass[letterpaper, 10 pt, conference]{ieeeconf}  

\IEEEoverridecommandlockouts                              

\usepackage{color}
\usepackage{graphics} 
\usepackage{epsfig} 
\usepackage{times} 
\usepackage{amsmath} 
\usepackage{amssymb}  
\usepackage{multirow}

\usepackage{array}
\usepackage{derivative}
\usepackage{graphicx}
\usepackage{pdflscape}  
\usepackage{xcolor}     
\usepackage{comment}
\usepackage[linesnumbered,ruled,vlined]{algorithm2e}
\usepackage{url}

\usepackage[linesnumbered,ruled,vlined]{algorithm2e}
\title{\LARGE \bf
Enhanced Detection Classification via Clustering SVM for Various Robot Collaboration Task
}

\author{Rui Liu$^{1}$, Xuanzhen Xu$^2$, Yuwei Shen$^3$, Armando Zhu$^4$, Chang Yu$^5$, Tianjian Chen$^6$, Ye Zhang$^*$
\thanks{$^{1}$Rui Liu be with Illinois Institute of Technology, IL 60616, USA {\tt\small { liuruiabc1}@gmail.com}}
\thanks{$^2$Xuanzhen Xu be with Snap Inc. at Seattle, WA 98121, USA {\tt\small {xuanzhenxu}@gmail.com}}
\thanks{$^3$Yuwei Shen is an Independent Researcher, USA {\tt\small {yuwei\_shen}@berkeley.edu}}
\thanks{$^4$Armando Zhu be with Carnegie Mellon University at Pittsburgh, PA 15213, USA {\tt\small {armandozhu}@cmu.edu}}
\thanks{$^5$Chang Yu be with Northeastern University at Boston, MA 02115, USA {\tt\small {chang.yu}@northeastern.edu}}
\thanks{$^6$Tianjian Chen be with The George Washington University at Washington, DC 20052, USA {\tt\small {chentianjian}@gwu.edu}}
\thanks{$^{*}$Ye Zhang be with University of Pittsburgh at Pittsburgh, PA 15213, USA {\tt\small {yez12}@pitt.edu}}
}
\begin{document}

\maketitle


\begin{abstract}

We introduce an advanced, swift pattern recognition strategy for various multiple robotics during curve negotiation. This method, leveraging a sophisticated k-means clustering-enhanced Support Vector Machine algorithm, distinctly categorizes robotics into flying or mobile robots. Initially, the paradigm considers robot locations and features as quintessential parameters indicative of divergent robot patterns. Subsequently, employing the k-means clustering technique facilitates the efficient segregation and consolidation of robotic data, significantly optimizing the support vector delineation process and expediting the recognition phase. Following this preparatory phase, the SVM methodology is adeptly applied to construct a discriminative hyperplane, enabling precise classification and prognostication of the robot category. To substantiate the efficacy and superiority of the k-means framework over traditional SVM approaches, a rigorous cross-validation experiment was orchestrated, evidencing the former's enhanced performance in robot group classification.

\end{abstract}

\section{Introduction}

In pursuit of classification for various kinds of robotics systems during a collaborative task that dynamically parameters to align with the center control platform requirements and underpins the foundational control mechanisms for sophisticated robot dynamics technologies~\cite{zhou2024machine,pan2023navigating,chen2024comprehensive,zhou2023distributed}, the accurate identification and forecasting of robot's group motion and distribution emerge as critical undertakings. Such endeavors aim not only to enhance collaboration efficiency and minimize emissions through tailored control strategies but also to integrate specific factors of robots into the design of robot dynamics and assistance systems~\cite{li2024ddn,zhang2020manipulator}. This necessitates a comprehensive analysis of robotics, encompassing both their distribution and feature conditions, as well as their status patterns (identifiable as functional or malfunction). Recent studies have significantly focused on the precise recognition and prediction of these robotics attributes, laying the groundwork for collaboration in various kinds of multi-robot systems.
\begin{figure}[thpb]
\centering
\includegraphics[scale=0.34]{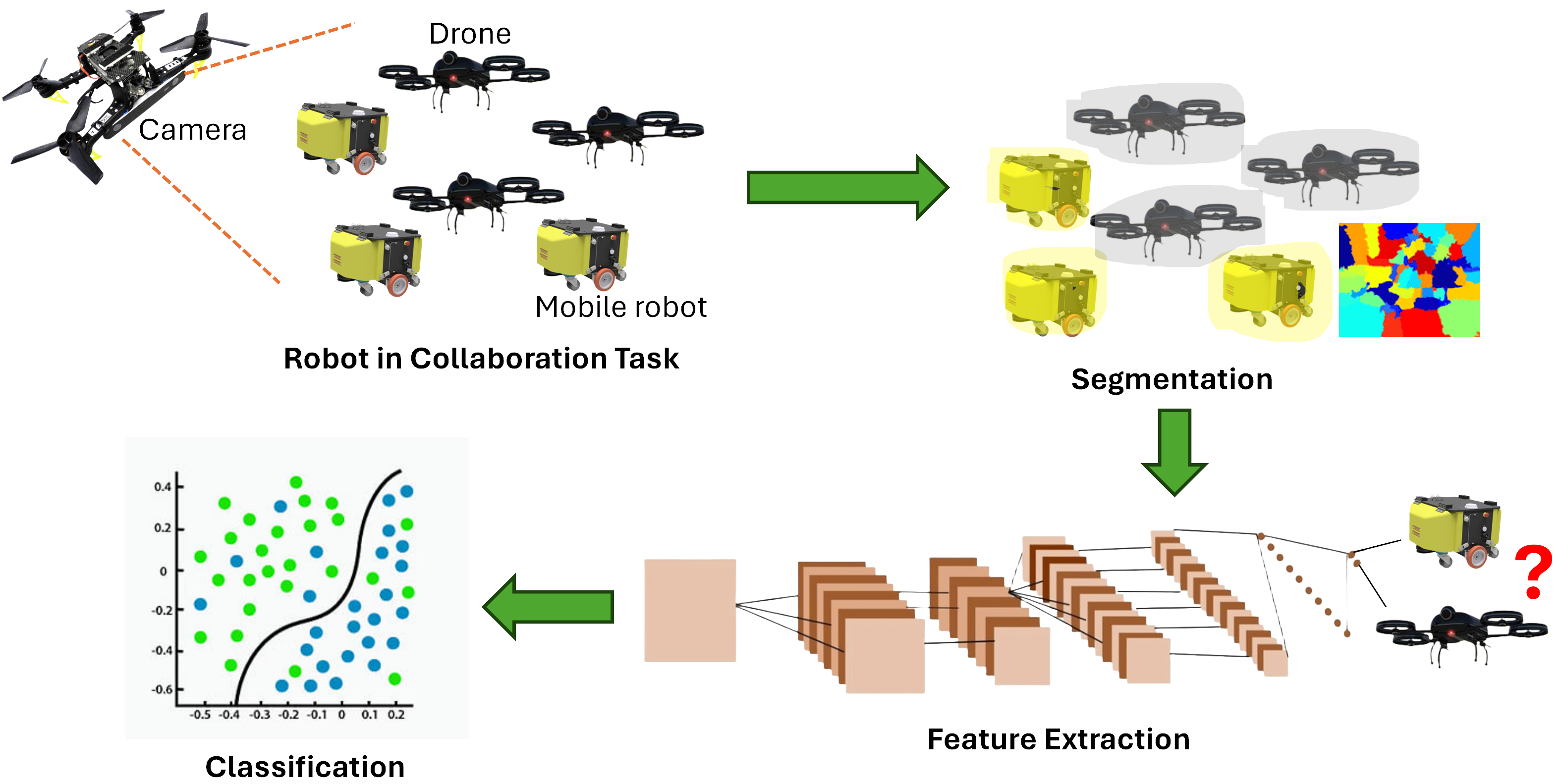}
\caption{Two different robot groups in a collaboration task; one flying robot installed a camera module as a ``commander" to detect and recognize the whole scenario of both groups to coordinate the collaboration.}
\label{figure0}
\end{figure}

Central to the recognition of robotics is the deployment of a model-based or indirect methodology. This approach necessitates the formulation of a comprehensive coordination encapsulating fundamental multi-agent collaboration such as perception~\cite{lauri2020multi}, manipulation~\cite{gao2023autonomous}, and obstacle navigation~\cite{pandey2017mobile}. Following the establishment of this model, robotics characteristics are delineated and analyzed. The utilization of stochastic process theories, particularly the Hidden Markov Model—a form of dynamic Bayesian network—has proven instrumental in elucidating the intricate correlations between observed data and underlying robot states, thereby facilitating the modeling and anticipation of rootics behaviors and decisions, especially in complex scenarios like intersection navigation. Moreover, advanced probabilistic models, including the ARX framework and its stochastic variant~\cite{zou2023joint}, the SS-ARX model, have been developed to accurately reflect the unpredictability inherent in robotics status, enabling refined classification and prediction of robotics distribution. 

This study introduces an expedited, direct methodology for identifying robot styles through the integration of k-means clustering and Support Vector Machine (SVM) techniques. Traditional classification methods such as SVM, ANNs, and ARX, despite their efficacy, often suffer from protracted computational times, particularly with non-linearly separable datasets~\cite{zhou2023semantic}. Our proposed k-SVM approach aims to surmount these limitations by streamlining the recognition process in three key phases: First, the clustering phase utilizes k-means to segment robotics feature data into distinct groups, thereby enhancing feature discrimination and reducing support vector quantities. Second, in the training phase, a discriminative hyperplane is constructed via the k-SVM method, enabling precise categorization of robot styles. Finally, experimental validation is undertaken through cross-validation techniques to affirm the method's efficiency and accuracy in robot style identification, showcasing the advantages of this novel approach.

\section{Pattern Recognition Method}
In this section, the parameters selected for the SVM and k-SVM are discussed. 


\subsection{Support Vector Machines}
In this investigation, we address the challenge of classifying robot patterns that are inherently non-linearly separable, manifesting as overlapping class distributions within a $q$-dimensional space~\cite{wang2024jointly}. A fundamental advantage of Support Vector Machines (SVM) utilized here, as highlighted by \cite{pisner2020support}, is their capacity to solve convex optimization problems, thereby ensuring a global optimum solution~\cite{deng2024compact}. The dataset comprises $n$ instances, each represented by a feature vector $\boldsymbol{\eta}_{i}$, associated with a binary target value $y_{i} \in \{-1,1\}$. Our objective is to delineate a hyperplane within a high-dimensional space that can discern between two distinct classes of robotics status. To accommodate the non-linear separability, we introduce slack variables $\xi_{i} \geqslant 0$, for each training case, enabling the formulation of a robust classification model that navigates the intricacies of overlapping distributions effectively~\cite{zhuang2022deperturbation}. For the new input data $\boldsymbol{\theta}$, its target value can be calculated by 

\begin{equation}
f(\boldsymbol{\theta})=\mathbf{w}^{\top}\Phi(\boldsymbol{\theta})+\boldsymbol{b} +\boldsymbol{\varepsilon}
\end{equation}
In the proposed model, $\Phi(\boldsymbol{\theta})$ represents a transformation into the feature space, while $\boldsymbol{b}$ serves as a bias vector, and $\boldsymbol{\varepsilon}$ denotes the slack variable. The formulation of the objective function is articulated as follows:

\begin{equation}
\min M\sum_{k=1}^{n}\boldsymbol{\varepsilon}_{k}+\frac{1}{2}\parallel \mathbf{w} \parallel ^{2}
\label{eq2}
\end{equation}
Where $M$ strategically modulates the balance between the penalty attributed to the slack variables and the optimization of the margin. Consequently, Equation (\ref{eq2}) is reformulated into its corresponding Lagrangian expression:

\begin{equation}
\mathcal{L}(\boldsymbol{\varsigma})=\sum_{j=1}^{n}\varsigma_{i}-\frac{1}{2}\sum_{j=1}^{n}\sum_{k=1}^{n}\boldsymbol{k} \cdot \varsigma_{j}\varsigma_{k}f_{j}f_{k} 
\end{equation}
In the derived formulation, $\boldsymbol{\varsigma} = {\varsigma_{j}}$ is designated as the set of Lagrangian multipliers, and $\boldsymbol{k}$ corresponds to the kernel function~\cite{zhibin2019labeled}, parameterized relative to its kernel parameters~\cite{sun2024transtarec}. The kernel function, inherently positive definite(P.D.), is exemplified specifically through the selection of a Gaussian kernel, which is delineated as follows:
\begin{equation}
\boldsymbol{k} = -\dfrac{1}{2} \alpha \parallel \boldsymbol{\nu}_{j} - \boldsymbol{\nu}_{k} \parallel ^{2} \\
\end{equation}

In this study, cross-validation and grid-search techniques are employed to ascertain the optimal parameters ($ M^*,\alpha^* $). Referencing, sequences of $M$ and $\alpha$ that exhibit exponential growth yield superior results in the identification of optimal parameters\cite{lin2024optimizing}. For the training of SVM, the initial values are set to ($ M_{0,+}, \alpha_{0,+}$) is set as ($ 2^{0},2^{-1} $). Subsequent to the dataset training, the optimal parameters are determined as $M^* = 2^{7}$ and $\alpha^* = 2^{-9}$. Utilizing these parameters, robot patterns are analyzed via the $k$-SVM method, culminating in the generation of the optimal separating hyperplane, as depicted in Fig. \ref{figure5} and Fig. \ref{figure5-1}.

\subsection{k-means Clustering}
Consider a dataset $\mathcal{D}$  comprising  $k$  training examples. Typically, these robot data sets include thousands of overlapping data points ( $\mathbf{x}_{i}, f_{i}$). To enhance the efficiency of the analytical model by reducing the number of support vectors and delineating raw feature parameters across diverse robot patterns~\cite{shen2024localization}, the  $k$-SVM method is utilized.

k-means is used to partition the raw data sets $ \{\mathbf{x}_{i},f_{i} \} $ into $ K $ ($ K \leqslant n $) clusters, forming a set $ \mathcal{W}:=\{(\hat{\mathbf{x}}_{k},f_{k}),\,k=1,2,\dots,N\} $. The $ (\hat{\mathbf{x}}_{k},f_{k}) $ is the subset of set $ \mathcal{W} $. In this paper, the k-means is calculated by optimizing the following objective function:

\begin{equation}
\arg\,\min_{\mathcal{W}} \sum_{k}^{U}\sum_{\mathbf{x}}\parallel \mathbf{x}-(\boldsymbol{\nu}_{k},f_{k}) \parallel ^{2}
\end{equation}
where $ (\boldsymbol{\nu}_{k},f_{k}) $ is the mean of point in set $ (\tilde{\mathbf{x}}_{k},f_{k}) $.

\subsection{Training Analysis}
Two typical collaboration tasks are discussed in the training results.

\begin{figure}[thpb]
\centering
\includegraphics[scale=0.4]{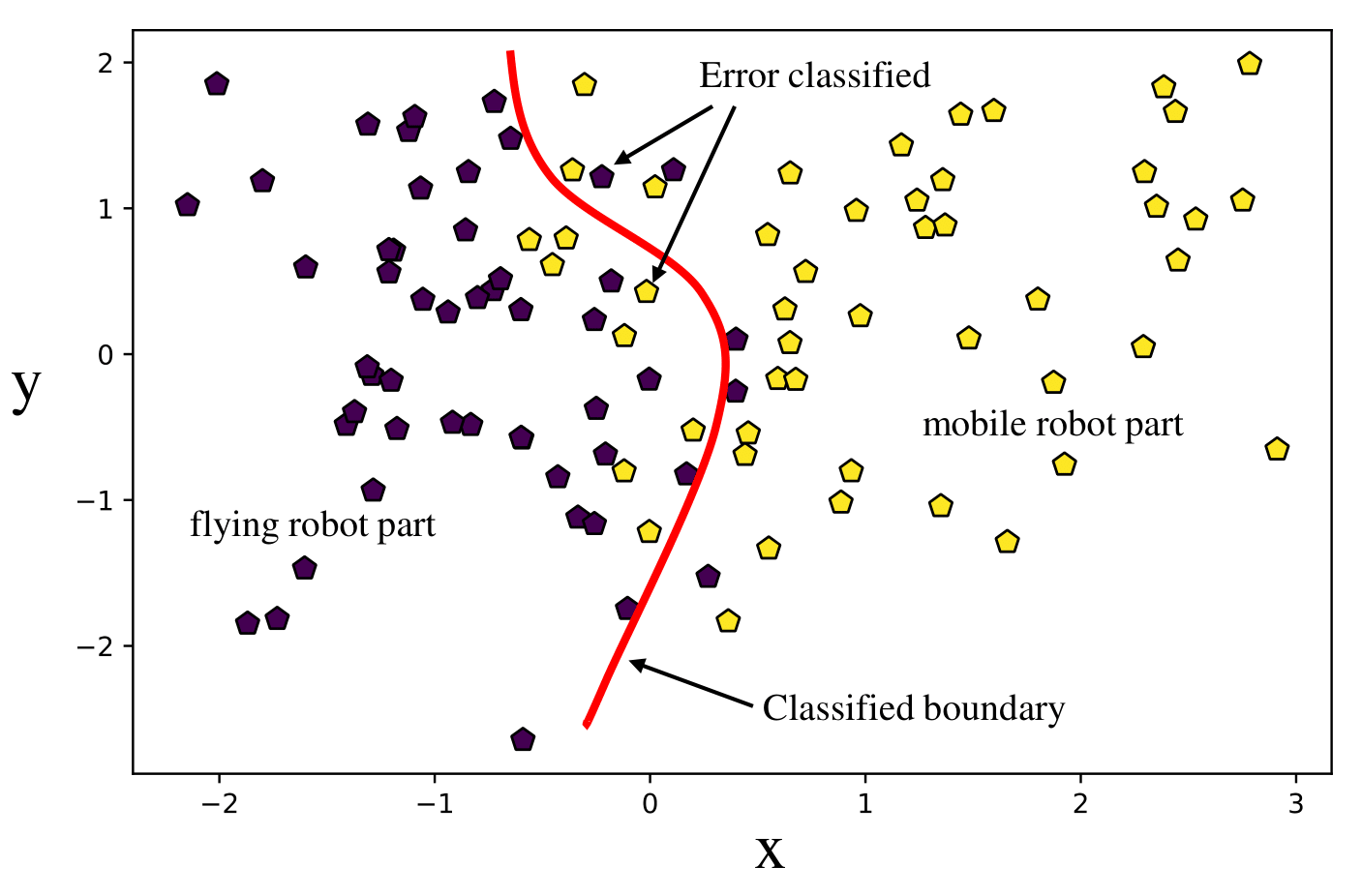}
\caption{The classifying results using k-SVM method for both kinds of robot. Yellow pentagon: mobile robot; black pentagon: flying robot.}
\label{figure5}
\end{figure}

\begin{figure}[thpb]
\centering
\includegraphics[scale=0.4]{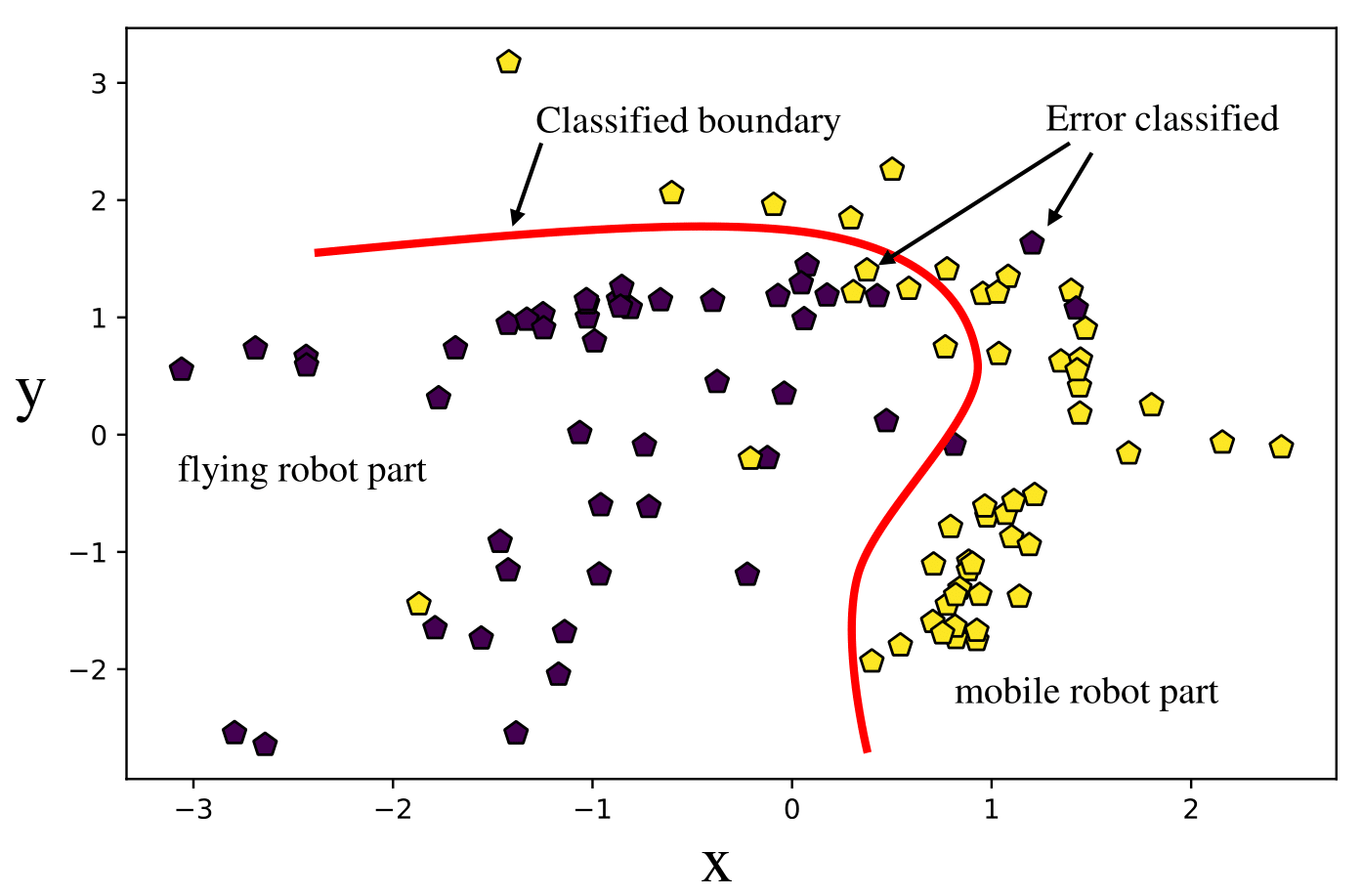}
\caption{The classifying results using k-SVM method for both kinds of robots. Yellow pentagon: mobile robot; black pentagon: flying robot.}
\label{figure5-1}
\end{figure}

For \textit{Task-1} in Fig. \ref{figure5}, a consortium of flying robots synergistically amalgamates with their terrestrial counterparts to fortify a strategic encirclement of a designated objective. Comprising nearly equivalent numbers, these aerial automatons not only contribute to the direct confrontation of the target but also augment the situational awareness of the ground-based units by conducting reconnaissance and surveillance of the surrounding milieu~\cite{zou2022unified}. The dual functionality facilitates an optimized distribution of tasks, where the flying units’ elevated vantage point provides a pivotal advantage in the real-time assessment of environmental dynamics~\cite{zhuang2022does}, thereby enhancing the operational efficacy and adaptive response capabilities of the mobile robots. Such integrative operations exemplify cutting-edge robotic synergy, showcasing an intricate ballet of aerial and terrestrial coordination aimed at achieving a fortified and comprehensive engagement with the target. According to Fig \ref{figure5}, the exposition delineated herein illustrates the classification paradigm of the k-SVM employed to discern between aerial and terrestrial robotic entities within a two-dimensional scenario~\cite{xiong2024tilp,liu2024deep,zi2024research}. This plot manifests two distinct clusters, each representing a class: one for flying robots and another for mobile robots. The decision boundary, ingeniously crafted by the SVM, is conspicuously demarcated with a hyperplane that optimally separates these two classes with maximal margin~\cite{zou2023multidimensional}, showcasing the quintessential embodiment of the k-SVM capability to enhance class separability~\cite{ru2022bounded}.
\begin{algorithm}[t]
\DontPrintSemicolon
\SetKwComment{Comment}{}{}
\KwData{train set $\boldsymbol{W}$,
test set $\Tilde{\boldsymbol{W}}$,
dictionary $\psi$,
iterations N
}
\KwResult{final dictionary $\psi$ and k-SVM model $(\omega, \eta, \xi, \lambda)$}
\BlankLine
\emph{Training Procedure }\\
Representation: apply k-means to obtain $X^0$ and its support $\Omega$\\
Support vectors: model $(\omega^0, \eta^{0}, \xi^{0},\lambda^{0})$ based on $X^0$ \\
\For{$k \in \{1,\dots,N \}$}{
DL: inner routine that learns $\psi^{k}$ and $X^{k}$ with fixed $\omega^{k-1}, \lambda^{k-1}$ \\
Support vectors: model $(\omega^{k}, \eta^{k}, \xi^{k}, \lambda^{k})$ based on fixed $X^{k}$ \\
}
\BlankLine
Representation: inner routine that obtains $\Tilde{X}$ from $\Tilde{\boldsymbol{W}}$\\
\caption{k-SVM Classfication Algorithem}
\label{alg:general_scheme}
\end{algorithm}
For \textit{Task-2} in Fig. \ref{figure5-1}, a contingent of flying robots is seamlessly integrated with a cadre of terrestrial robots to collectively defend a designated target. The scenario necessitates a strategic adaptation where both contingents—comprised of closely matched numbers—employ a cross-defense pattern to enhance the robustness of their protective strategy. The flying robots not only participate directly in the defense but also provide crucial reconnaissance~\cite{deng2023prosgnerf}, leveraging their elevated perspectives to monitor environmental conditions and potential hazards surrounding the target. Fig. \ref{figure5-1} shows the classification of these robots exemplifies a sophisticated application of cross-modal collaboration.

\section{Experiments in Robot Classfication}
In this section, the robot distribution and the classification of the training data are discussed. 

\subsection{Training-Data classification in Experiments}
In our empirical investigation, we meticulously curate a dataset to train classifiers, specifically k-SVM and traditional SVM, for the effective differentiation of two distinct groups of robots: aerial and terrestrial. The aggregate count of entities, denoted as $N$, is stratified into three diverse subsets, each embodying a unique distribution paradigm to enrich the classifier's training phase~\cite{liu2024deep}. This methodological diversification is designed to robustly test the resilience and adaptability of the classifiers under varying conditions reflective of real-world operational environments. By manipulating the distribution characteristics within each subset—varying densities and spatial configurations—we facilitate a comprehensive evaluation of the classifiers' performance in accurately mapping and distinguishing between the aerial and mobile robot cohorts on a conceptual representation of the operational arena~\cite{zhang2024deepgi}. This rigorous approach not only enhances the generalizability of our findings but also ensures that the classification algorithms can effectively handle diverse and dynamically changing scenarios, thereby bolstering their applicability in practical deployment settings where robustness and precision are paramount~\cite{yao2023ndc}.

\section{Classification Performance Evaluation}

The assessment method for the proposed classfier, i.e., k-SVM, is described in this section and the testing results are presented and discussed.

\subsection{Evaluation Method}

To ascertain the efficacy of the proposed classification methodology, the robust evaluation mechanism known as \textit{cross-validation} is employed. The dataset comprises recordings from two subsets of robots: mobile robots, numbered at $r$, and flying robots, with the remainder $N-r$. Performance evaluation involves the segmentation of the dataset into  $\eta$  ($\eta < N$), $\eta \in \mathbb{N}_+$ arbitrary subsets, with the classifier being trained on $N-1$ subsets and validated on the remaining subset~\cite{zou2023capacity}. This technique is designated as cross validation, with $\eta$ specifically set to $1$. The precision of the robot-pattern recognition system is subsequently quantified.

\begin{itemize}
\item For the flying robot:

\begin{equation}
\eta_{f} = \frac{K_{f,m}}{\sum K_{\sum,f}}, \quad 0\leqslant\lambda_{f} \leqslant 1
\end{equation}

\item For the mobile robot:

\begin{equation}
\eta_{m} = \frac{K_{f,m}}{\sum K_{\sum,m}}, \quad 0 \leqslant \lambda_{m} \leqslant 1
\end{equation}

\end{itemize}
where $K$ signifies the associated count of robot patterns. Specifically, $ K_{f,m} $ represents the quantity of clustering points accurately categorized within the pattern region of the flying robot.

\subsection{Testing Results and Analysis}

For different testing data sets $\boldsymbol{W}$, the test results corresponding to training data set $ \boldsymbol{W} $ are conducted using the off-line and online methods. Then, the comparisons between k-SVM and SVM are presented.

\begin{figure}[thpb]
\centering
\includegraphics[scale=0.4]{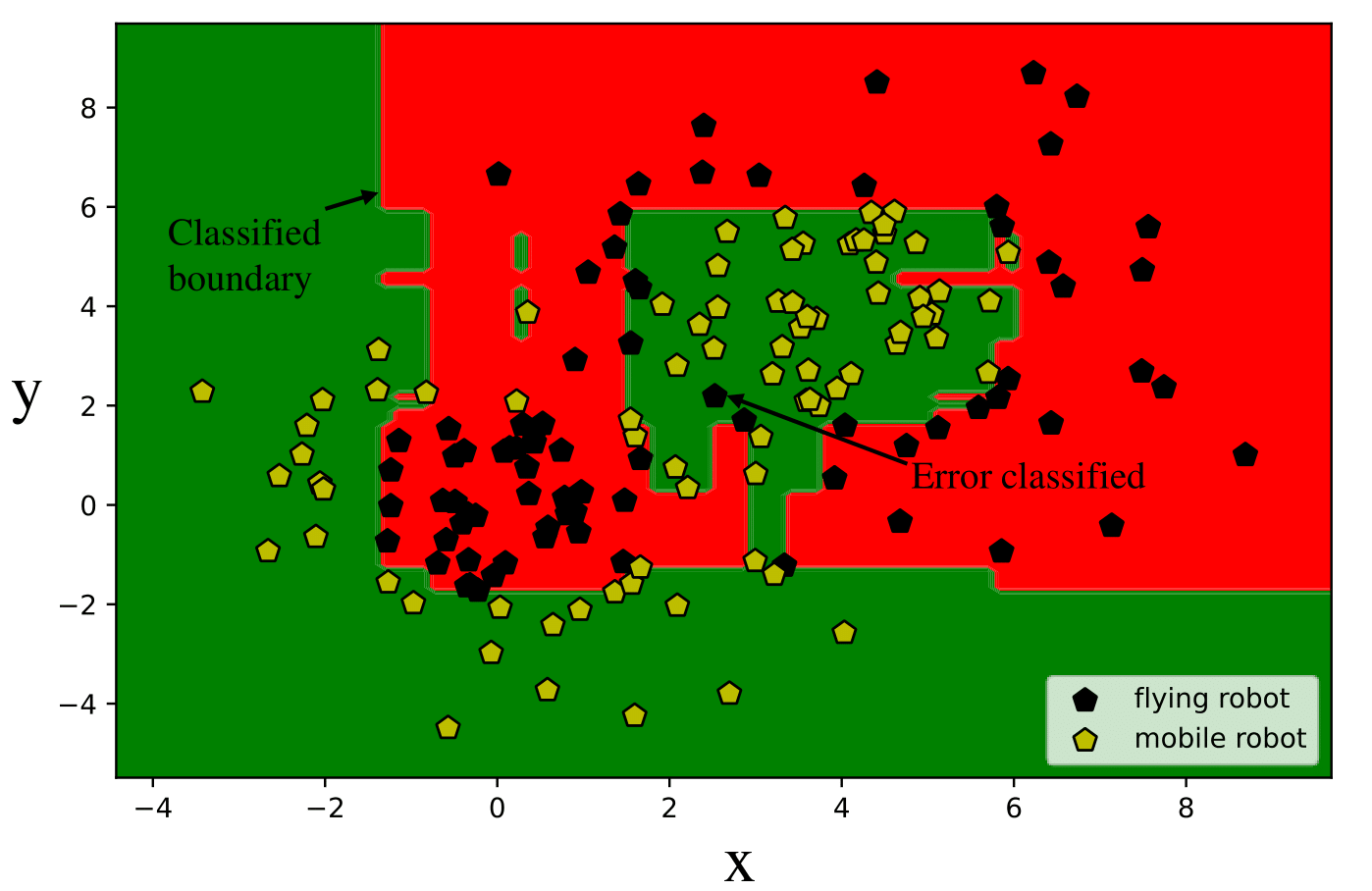}
\caption{The testing evaluation for two types robot using the K-SVM method based on training data in Fig. \ref{figure5}.}
\label{fig4}
\end{figure}

\begin{figure}[thpb]
\centering
\includegraphics[scale=0.4]{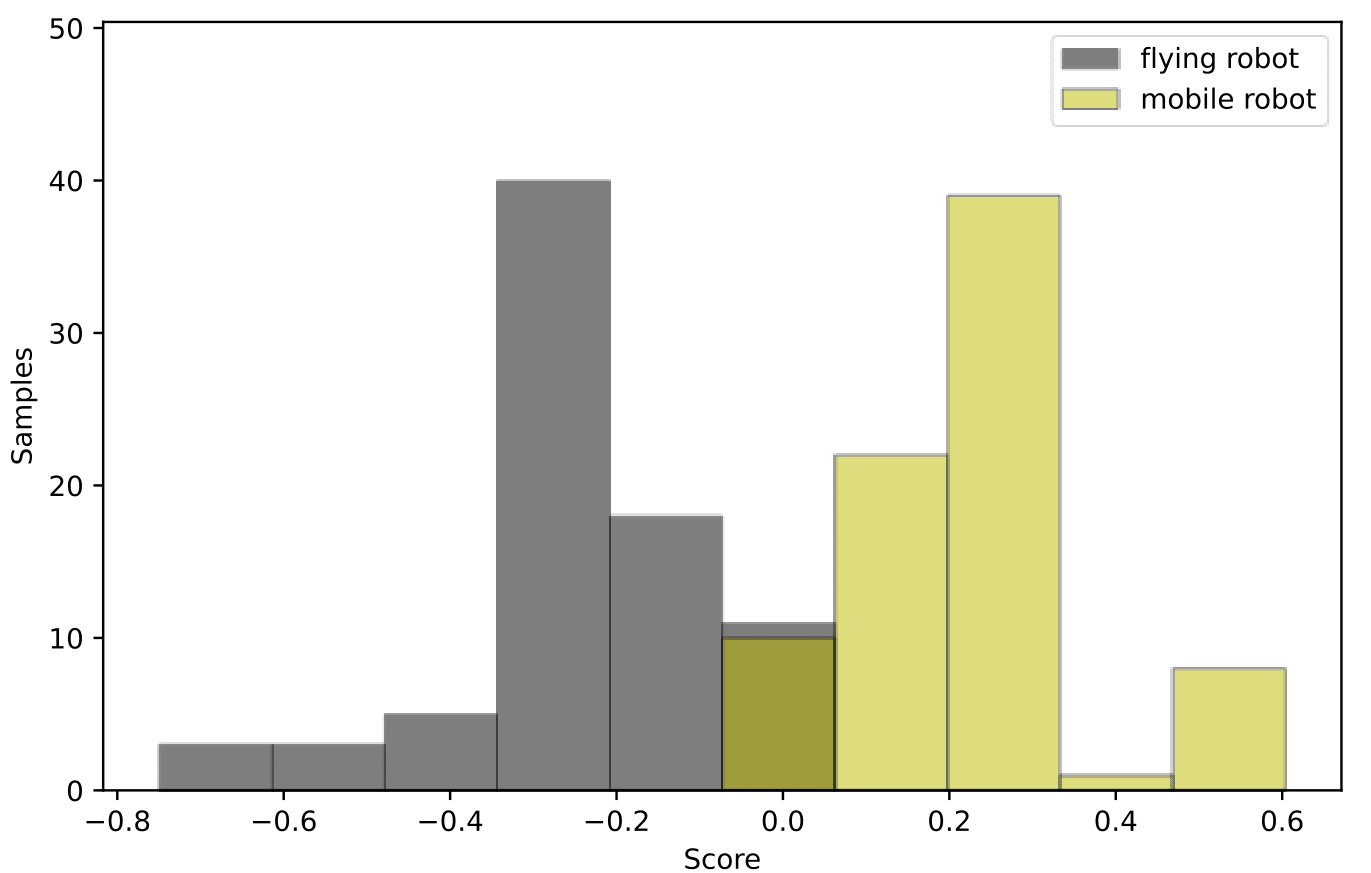}
\caption{Classification results for two types robot using k-SVM method.}
\label{fig-5}
\end{figure}

We assume that both kinds of robotics patterns are treated as constant in a fixed interval~\cite{yao2024building}. In this paper, the past information or robot pattern of the robot during a fixed time span $ \delta $ is adopted to represent the current robot pattern. Therefore, the error of classification is described as follows:

\begin{equation}
e_{s}=\Re(\boldsymbol{w}_{s-\delta,p},y_{s-\delta})
\end{equation}
In our refined analysis, the classification performance of k-SVM is contrasted with traditional SVM methods across three experimental conditions distinguished by the population of robots, $N$, segmented into cohorts of 100, 200, 300, and 400, each subjected to unique distribution patterns. This systematic approach ensures a robust evaluation under varied yet controlled conditions~\cite{yao2023improving}, simulating potential real-world operational settings. The temporal clustering of robot features within the period $[0, \dots K]$, where each point $\boldsymbol{w}_{s-\delta}$ represents a clustering centroid calculated over a time span $\delta=0.12$ , allows for the generation of K-clustered datasets. The efficacy of these classifications is substantiated through comprehensive results presented in Figures \ref{fig4} and \ref{fig-5}, and Table \ref{tb-1}. It is evident from Table \ref{tb-1} that k-SVM outperforms traditional SVM in recognizing diverse robot patterns, achieving lower classification errors as the dataset size increases~\cite{zhang2024development}. This inverse correlation between training volume and error rate underscores the enhanced adaptability and accuracy of the k-SVM approach in handling complex data distributions.


\begin{table}[ht]
\begin{center}
\begin{tabular}{|c|c|c|c|c|}
\hline
& \multicolumn{4}{c|}{Normal distribution of Robots} \\ \hline
& $N = 100$ & $N = 200$ & $N = 300$ & $N = 400$ \\ \hline
SVM & 0.2941 & 0.1259 & 0.3214 & 0.1561 \\ 
k-SVM & \textbf{0.1542} & \textbf{0.1080} & \textbf{0.2955} & \textbf{0.0996} \\  \hline\hline
& \multicolumn{4}{c|}{Uniform distribution of Robots (2 degrees of freedom)} \\ 
\hline
& $N = 100$ & $N = 200$ & $N = 300$ & $N = 400$ \\ \hline
SVM & 0.2302 & 0.1581 & 0.1662 & 0.0912 \\ 
k-SVM & \textbf{0.2966} & \textbf{0.1710} & \textbf{0.6861} & \textbf{0.3322} \\  \hline\hline
& \multicolumn{4}{c|}{Poisson distribution of Robots (4 degrees of freedom)} \\ 
\hline
& $N = 100$ & $N = 200$ & $N = 300$ & $N = 400$ \\ \hline
SVM & 0.3189 & 0.2816 & 0.2104 & 0.1549 \\ 
k-SVM & \textbf{0.2714} & \textbf{0.2046} & \textbf{0.1067} & \textbf{0.0851} \\ \hline
\end{tabular}
\end{center}
\caption{Comparison of the classification error of k-SVM and SVM in different distributions of robots.}
\label{tb-1}
\end{table}

\section{Conclusion}
A rapid pattern-recognition method, called k-SVM, is developed by combining the k-means clustering and SVM and subsequently applied to recognize the robot's curve-negotiating patterns, i.e., aggressive and moderate. The k-SVM, compared with SVM, can not only shorten the recognition time but also improve the recognition for classification issues that are not linearly separable. First, to reduce the number of support vectors, the $ k $-means clustering method is applied, clustering the original data sets into $ K $ subsets. And then, based on the clustering results, the SVM is applied to generate the hyperplane for datasets with different labels. Last, the cross-validation experiments are designed to show the benefits of the proposed method. The testing results show that the k-SVM is able to not only shorten the training time for the classification model but also improve recognition, compared with the SVM method. 






\bibliographystyle{ieeetr}
\bibliography{ref}

\end{document}